\pdfoutput=1
\documentclass[letterpaper, 10 pt, conference]{ieeeconf}  % Comment this line out if you need a4paper

\IEEEoverridecommandlockouts                              % This command is only needed if 
\usepackage{cite}
\usepackage{amsmath,amssymb,amsfonts}
\usepackage{algorithmic}
\usepackage{graphicx}
\usepackage{textcomp}
\usepackage{xcolor}
\usepackage{lipsum}
\usepackage{bm}
\usepackage{float}
\DeclareTextSymbolDefault{\DH}{T1}

\usepackage[utf8]{inputenc}
\usepackage{pgfplots}
\DeclareUnicodeCharacter{2212}{−}
\usepgfplotslibrary{groupplots,dateplot}
\usetikzlibrary{patterns,shapes.arrows}
\pgfplotsset{compat=newest}
\usepackage{subfigure}
\title{\LARGE \bf
A Time and Place to Land: Online Learning-Based Distributed MPC for Multirotor Landing on Surface Vessel in Waves
}

\author{Jess Stephenson, William S. Stewart, and Melissa Greeff % <-this % stops a space
\thanks{The authors are with Robora Lab (www.roboralab.com), Queen's University; and affiliated with Ingenuity Labs Research Institute. E-mails: jess.stephenson@queensu.ca, william.stewart@queensu.ca, melissa.greeff@queensu.ca.}%
}

\begin{document}

\maketitle
\thispagestyle{empty}
\pagestyle{empty}

%%%%%%%%%%%%%%%%%%%%%%%%%%%%%%%%%%%%%%%%%%%%%%%%%%%%%%%%%%%%%%%%%%%%%%%%%%%%%%%%
\begin{abstract}
Landing a multirotor unmanned aerial vehicle (UAV) on an uncrewed surface vessel (USV) extends the operational range and offers recharging capabilities for maritime and limnology applications, such as search-and-rescue and environmental monitoring. However, autonomous UAV landings on USVs are challenging due to the unpredictable tilt and motion of the vessel caused by waves. This movement introduces spatial and temporal uncertainties, complicating safe, precise landings. Existing autonomous landing techniques on unmanned ground vehicles (UGVs) rely on shared state information, often causing time delays due to communication limits. This paper introduces a learning-based distributed Model Predictive Control (MPC) framework for autonomous UAV landings on USVs in wave-like conditions. Each vehicle's MPC optimizes for an artificial goal and input, sharing only the goal with the other vehicle. These goals are penalized by coupling and platform tilt costs, learned as a Gaussian Process (GP). We validate our framework in comprehensive indoor experiments using a custom-designed platform attached to a UGV to simulate USV tilting motion. Our approach achieves a 53\% increase in landing success compared to an approach that neglects the impact of tilt motion on landing.

\end{abstract}

%

%%%%%%%%%%%%%%%%%%%%%%%%%%%%%%%%%%%%%%%%%%%%%%%%%%%%%%%%%%%%%%%%%%%%%%%%%%%%%%%%
\section{INTRODUCTION}  %1-1.5 page
Multirotors are agile and provide aerial information but are limited by a shorter battery life. Uncrewed surface vessels (USVs) offer an extended range in marine and limnology applications but are constrained (e.g., view, speed). Teams of multirotors and USVs can enhance maritime applications, including search-and-rescue, surveillance, and remote monitoring \cite{wu2023cooperative}, \cite{vas2015}, \cite{wang2023cooperative}. For these remote applications, the ability to autonomously land a multirotor on a USV for recharging is critical.

A safe landing requires safe and reliable performance despite limited vehicle communication and temporary communication loss. Furthermore, when the USV encounters rough water conditions, the precise location and timing of the landing are crucial to prevent damage to the multirotor due to the USV's severe tilt at touchdown \cite{xia2022landing}.

\begin{figure}
\centering
\includegraphics[width=0.9\linewidth,trim={0cm 0cm 0cm 0cm},clip]{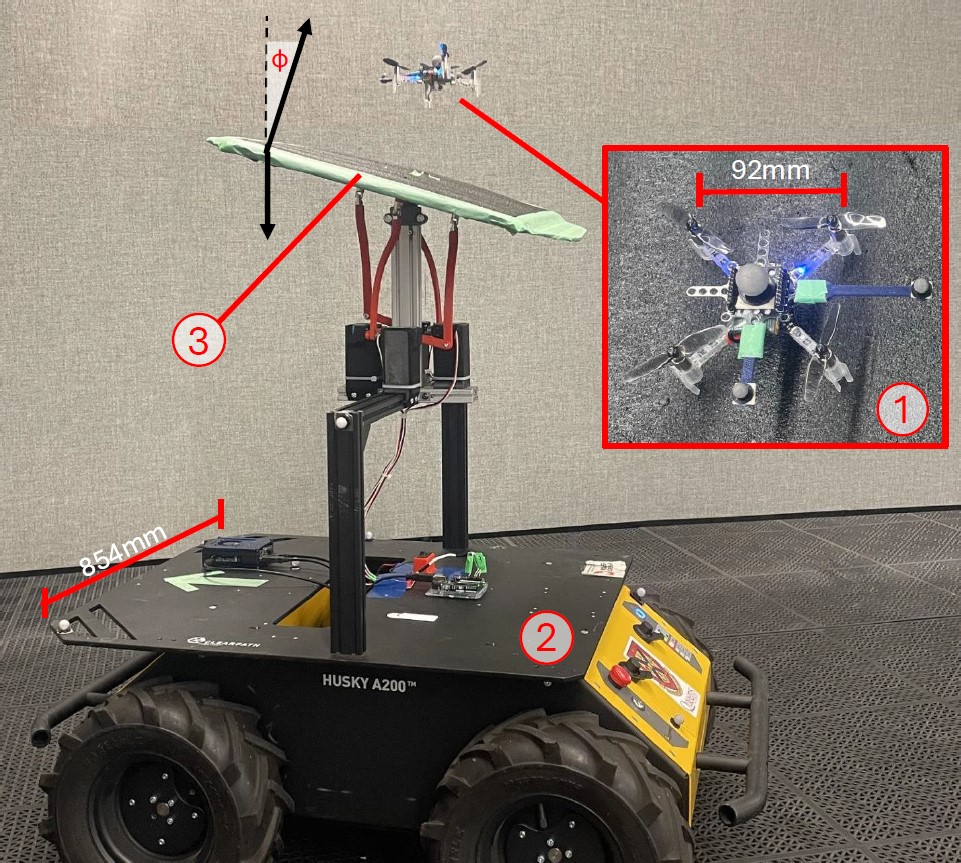}
\caption{Experimental setup showing (1) the Bitcraze Crazyflie 2.1 multirotor and (2) ClearPath Robotics Husky unmanned ground vehicle (UGV) equipped with (3) a custom-designed tilting platform. The multirotor communicates with the UGV for cooperative landing. The platform emulates the tilting motion of a USV in waves, enabling experimental validation of the proposed distributed model predictive control (MPC) framework.}
\label{F1}
 \vspace{-5mm}
\end{figure}

% Landing on ground vehicle 
Several studies have addressed the detection and motion estimation of moving platforms using a multirotor's onboard vision to enable its landing \cite{falanga2017}, \cite{keipour2022visual}, \cite{morando2024vision}. In this paper, we address a different problem: how to cooperatively coordinate the landing of a multirotor on a moving \textit{tilting} platform (representative of a USV) whose spatial-temporal tilt motion is only partially known. We, therefore, assume that state measurements are available for both the multirotor and our platform in Fig. \ref{F1}.   

Autonomous landing of a multirotor unmanned aerial vehicle (UAV) on a moving unmanned ground vehicle (UGV) without a tilted landing platform is commonly tackled through Model Predictive Control (MPC), e.g., \cite{mohammadi2020vision}, \cite{garegnani2021autonomous}, or sliding mode control (SMC), e.g., \cite{khaled2016}, \cite{modali2020sliding}. However, these approaches assume that the landing or rendezvous location is known beforehand or can be updated via an online heuristic. An alternative approach develops a distributed rendezvous control approach that drives the two vehicles' relative position errors and velocities to zero \cite{daly2011}. Given that the control signal applied to each vehicle requires knowledge of the state of the other vehicle and that the two systems are physically separated, there will be inherent time delays. Stability in the presence of these time delays has been analyzed. 

% coordinate a rendezvous for the two vehicles
Landing on a USV differs from landing on a UGV due to the USV's tilt motion \cite{novak2024uav_usv}. This significantly impacts the safe, optimal rendezvous location and landing time. Landing on a USV is commonly tackled through MPC, e.g., centralized \cite{persson2021model}, decentralized \cite{gupta2022landing}, or distributed MPC \cite{bereza2020distributed}. However, these methods either assume that the USV is stationary or that the landing location is specified and, therefore, known before the flight. In our work, the location and time of the landing is not known a priori. Building on our previous work in \cite{stephenson2024distributed}, we combine tracking and cooperative coordination into a distributed MPC by treating the landing problem as a consensus problem where both vehicles have different goal locations and iteratively update their states and goals until they achieve self-organized consensus \cite{kohler2024distributed}. 

% work on landing on a USV
A significant bottleneck to research on autonomous UAV-USV cooperative landing is the cost of surface vessels and the proximity to a suitable geographical testing location. Consequently, several approaches are limited to simulation, e.g., our previous work in \cite{stephenson2024distributed}. In this work, we first develop and build a low-cost testbed platform to mimic the spatial-temporal tilting motion of a USV in waves, see Fig \ref{F1}. This testbed allows us to demonstrate our proposed novel control framework in a controlled indoor environment and enables future benchmarking of alternative strategies (e.g., Reinforcement Learning (RL)-based approaches \cite{ramos2018}, \cite{wang2023vision}).

% Landing on a boat
A related approach explores landing a multirotor on a tilted platform mounted on a moving UGV \cite{Panagiotis2015}. This approach has two key differences from our work. Firstly, the tilted platform is fixed and does not rotate based on space and time (as a USV in waves does). Secondly, the UGV follows a prescribed trajectory without cooperating with the UAV. In our work, we build a tilting platform mounted on a UGV, see Fig. \ref{F1}, that adjusts its tilt according to a spatial-temporal model to mimic the motion of USV in waves. Furthermore, similar to \cite{daly2011}, we consider our testbed platform a cooperating vehicle whose trajectory is not known a priori. 

% Distributed MPC
Unlike our previous work in \cite{stephenson2024distributed}, which only showed simulation results for a known spatial-temporal wave model, in this paper, we demonstrate our MPC framework in multirotor experiments using our developed platform in Fig. \ref{husky} and learn the spatial-temporal wave model from data. The contributions of this paper are  threefold:
\begin{itemize}
    \item We present a novel learning-based distributed MPC strategy for
the cooperative landing of a multirotor on a USV by learning a spatial-temporal wave model as a Gaussian Process (GP).
\item We design and build a low-cost testbed platform mounted to a UGV that can mimic the spatial-temporal tilting motion of waves. 
\item We perform substantial experimental validation of our proposed distributed MPC framework for safe landing on a cooperative tilting platform. By mimicking harsh wave conditions with our platform, we achieve a 53\% landing success rate improvement over an approach that neglects the spatial-temporal tilt motion. 
\end{itemize}

\section{BACKGROUND}    %1.5 pages

\subsection{Model Predictive Control}
We consider a discrete-time dynamics model for our systems with continuous state $\mathbf{x}_i \in \mathcal{X}$ and input spaces $\mathbf{u}_i \in \mathcal{U}$. We denote the time-discretized evolution of the system $f: \mathcal{X} \times \mathcal{U} \rightarrow \mathcal{X}$ is:
\vspace{-2mm}
\begin{equation}
    \mathbf{x}_{i+1} = f(\mathbf{x}_i, \mathbf{u}_i),
    \label{eq_dyn}
\end{equation}
where the index $i$ refers to the states and inputs at time $t_i$. A common dynamics model for a USV is provided in \cite{fossen2011handbook}, while one can be found in \cite{nan2022nonlinear} for a multirotor UAV.
% state for UAV and USV

At each time step $i$, the MPC takes the current state $\mathbf{x}_{i} = \mathbf{x}_{\text{init}}$ and produces a sequence of optimal states $\mathbf{x}^{*}_{0:N}$ and control commands $\mathbf{u}^{*}_{0:N-1}$. The notation $*$ denotes the optimal solution, and $_{0:N}$ denotes the value for each time step from the current time step $i$ to $i+N$ where $N \in \mathbb{Z}$ is the prediction horizon. The sequence of optimal states and control commands can be computed by solving an optimization online using a multiple-shooting scheme \cite{Bock1984AMS}. The first control command is applied to the vehicle, and the optimization problem is solved again in the next state. Consider the setpoint $\mathbf{x}_g \in \mathcal{X}$, by minimizing a cost $J(\cdot)$ over a fixed time horizon $N$ at each control time step, MPC solves the constrained optimization as:
\begin{equation}
\begin{aligned}
	\min_{\mathbf{x}_{0:N}, \mathbf{u}_{0:N-1}} & \quad J(\mathbf{x}_{0:N}, \mathbf{u}_{0:N-1}, \mathbf{x}_g) \\
\textrm{s.t.} \quad & \mathbf{x}_{k+1} = f(\mathbf{x}_k, \mathbf{u}_k) \quad \forall k \in \mathcal{K} \\ %= 0,..., N-1\\
  &\mathbf{x}_{k} \in \mathcal{X} \quad \forall k \in \mathcal{K}  \\
  &\mathbf{u}_{k} \in \mathcal{U} \quad  \forall k \in \mathcal{K}  \\
  &\mathbf{x}_{0} = \mathbf{x}_{\text{init}},
\end{aligned}
\label{eq_mpc}
\end{equation}
where $\mathcal{K} := \mathbb{Z} \cap [0, N-1]$. The cost $J(\cdot) = J_{\textrm{track}}(\cdot)$ is commonly selected as quadratic in the tracking errors and inputs as:
\vspace{-2mm}
\begin{equation}
    J_{\textrm{track}}(\cdot) = \sum_{k=1}^{N} (\mathbf{{x}}_k - \mathbf{x}_g)^\intercal\mathbf{Q}( \mathbf{{x}}_k - \mathbf{x}_g) + {\mathbf{u}^\intercal_{k-1}} \mathbf{R} \mathbf{u}_{k-1},
    \label{eq_cost_track}
\end{equation}
where $\mathbf{Q} \succeq 0$ and $\mathbf{R} \succ 0$ are selected matrices that weight the position error (between the vehicle's position and its goal) and control effort, respectively.

\subsection{Gaussian Processes (GPs)}
GP regression is used to approximate a nonlinear function, $f_w(\mathbf{a}) : \mathbb{R}^{\text{dim}(\mathbf{a})} \rightarrow \mathbb{R}$, from input $\mathbf{a}$ to the function output. We use the notation $f_w(\mathbf{a})$ to be consistent with Sec. \ref{meth_gp}. As data points are collected, the set of possible functions for $f_w(\mathbf{a})$ is refined. Function values $f_w(\mathbf{a})$, that are evaluated at inputs $\mathbf{a}$, are random variables and the GP assumes that any finite number of these variables have a multivariate or joint Gaussian distribution. 

For this approach, we define a prior mean for the function $f_w(\mathbf{a})$, generally set to 0, and a kernel function $k(.,.)$. The kernel function computes and encodes the covariance or similarity between all pairs of datapoints. A common kernel function is the squared-exponential (SE) function:
\begin{equation}
k(\mathbf{a}, \mathbf{a}^\prime) = \sigma^2_\eta \exp ( -\frac{1}{2} (\mathbf{a} - \mathbf{a}^\prime)^\intercal \mathbf{M}^{-2} (\mathbf{a} - \mathbf{a}^\prime)) + \delta_{ij} \sigma_\omega^2,
\label{eq_se}
\end{equation}
where we define three hyperparamters, the prior variance $\sigma^2_\eta$, measurement noise $\sigma_\omega^2$, where $\delta_{ij}=1$ if $i=j$ and 0 otherwise, and the length scales which are the diagonal elements of the matrix $\mathbf{M}$. The length scales define how quickly the function value $f_w(\mathbf{a})$ changes with respect to $\mathbf{a}$. GP regression optimizes these hyperparameters by solving a maximum log-likelihood problem \cite{rasmussen2006gaussian}.

\section{METHODOLOGY}   %1-1.5 pages Jess
To facilitate safe landing coordination, we first learn a tilt model, representative of the spatial and temporal impact of waves on the tilt angle of a USV, using GP regression. We apply this model within a novel distributed MPC framework, adapted from \cite{stephenson2024distributed}. Finally, we build a custom tilting platform to validate our proposed framework in indoor experiments. 

%In this work we build on our distributed MPC control scheme as presented in \cite{stephenson2024distributed}. This previous work in simulation showed how the proposed scheme can coordinate a safe landing location for a UAV on a USV in harsh wave conditions. We now present experimental validation for multi-agent cooperative landing in harsh waves by leveraging our novel testbed platform. The tilting platform simulates the pitch and roll that a USV would experience in waves, while staying on dry land by being mounted on an unmanned ground vehicle (UGV). Furthermore, while our previous work assumed that the spatial-temporal wave model was known a priori, this paper takes an additional step towards full autonomy of the system by learning the wave model online using a Gaussian Process (GP).

\begin{figure*}[ht]
 \centering
 \subfigure[Full Testbed System]{
      \centering
      \label{husky_full}
      \includegraphics[width=0.3\textwidth]{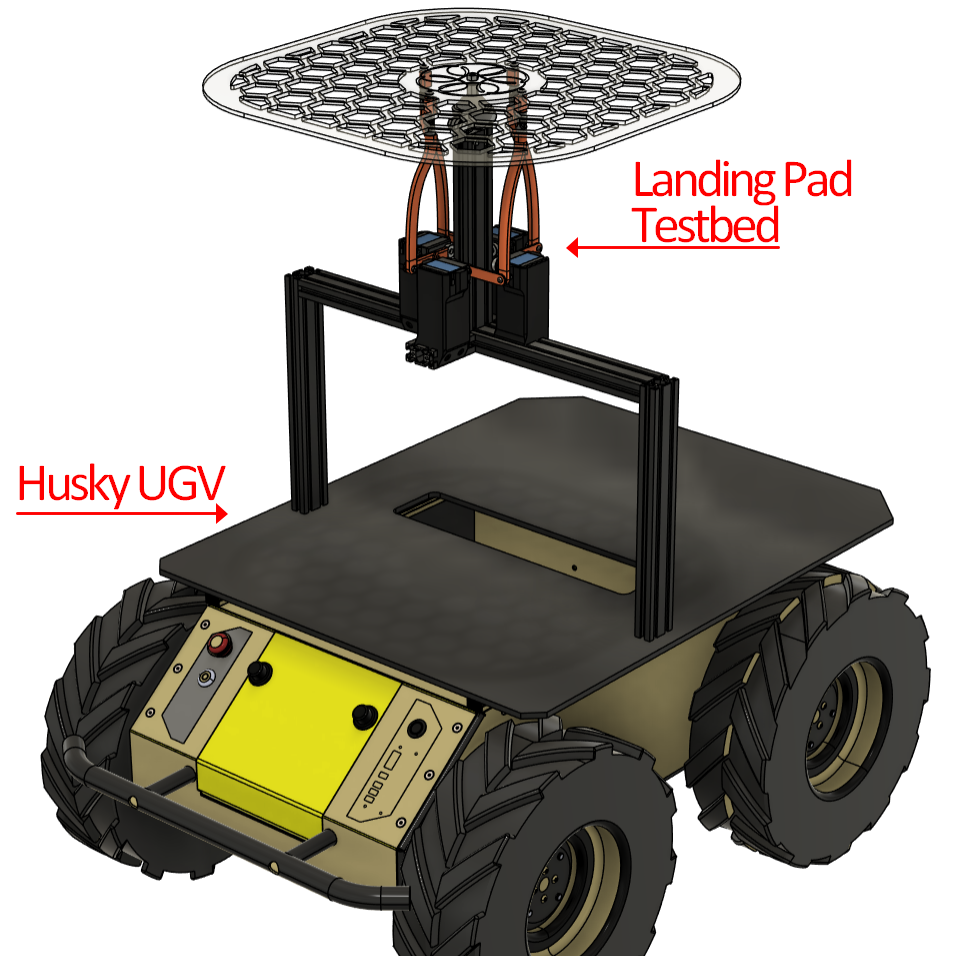}
		} 
 \subfigure[Landing Pad Configuration]{
      \centering
      \label{husky_apparatus}
      \includegraphics[width=0.3\textwidth]{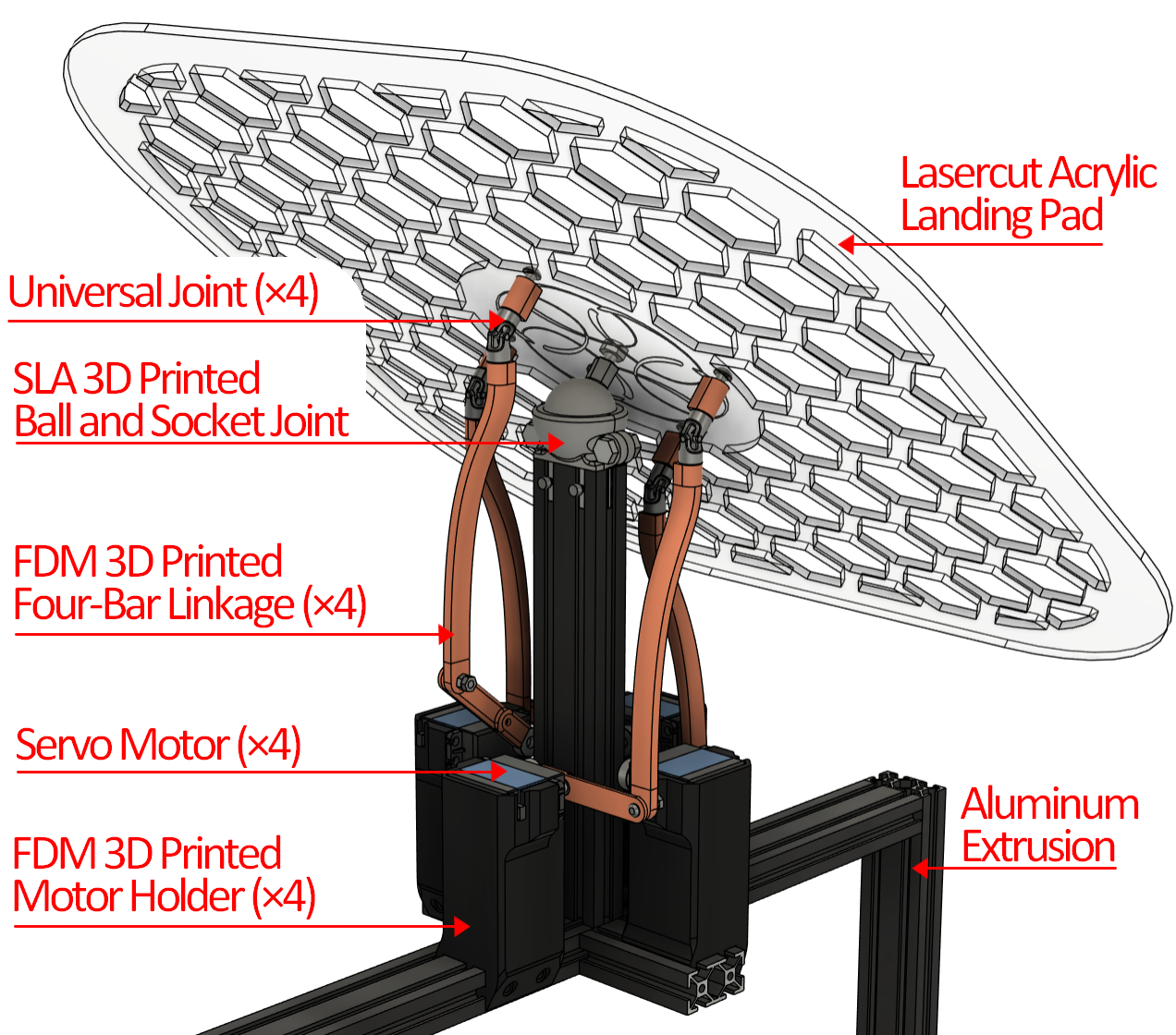}
		} 
\subfigure[Section View]{
      \centering
      \label{husky_section}
      \includegraphics[width=0.3\textwidth]{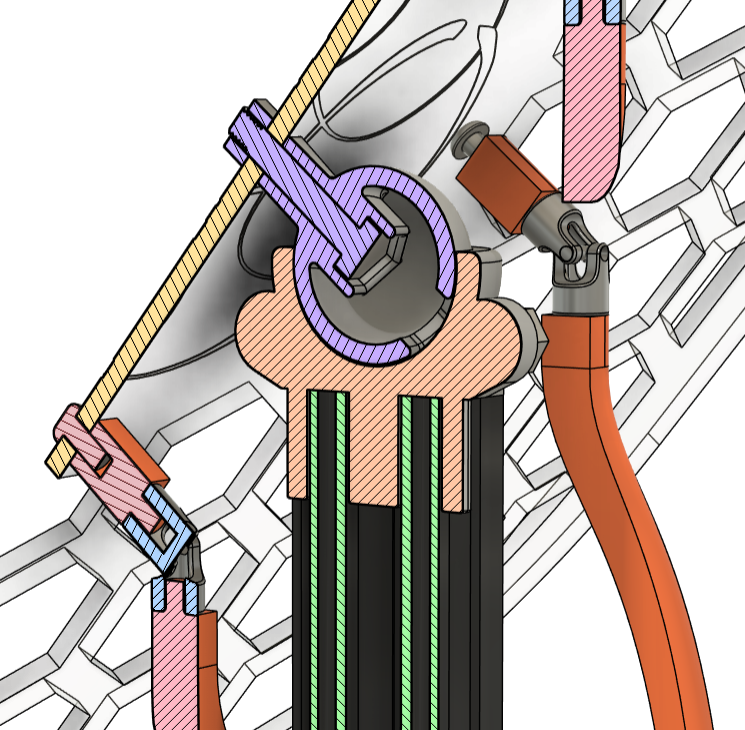}
		} 
 \caption{Overview of our custom tilting platform for ground vehicles. In (a), the platform is shown mounted on the deck of the ClearPath Robotics Husky, (b) shows a closer view of the platform's components, and (c) highlights the ball-and-socket joint and the linkage mechanism.
}

 \label{husky}
  \vspace{-5mm}
\end{figure*}

\subsection{Learning the Tilt Model as a Gaussian Process}
\label{meth_gp}
%move the mpc cost equations that contain mean and cov down here and include the generic fw in the background instead.
%From prev paper section C. spatial-temporal map of waves
%To learn the spatial-temporal map of waves in an area, the USV is used as a mobile sensing platform to make observations about the waves. We see in \cite{sears2023mapping} that GP regression can be used successfully to create a spatial-temporal wave model by leveraging hyperparameters to assume spatial and temporal correlations in the data. At any USV postion $x$ and $y$ and time $t$, inertial measurements can obtain an estimated pitch $\hat{\alpha}$ and estimated roll $\hat{\beta}$. The estimated wave tilt $\hat{\phi}$ at a spatial-temporal location is then given by:
%\begin{equation}
%	\phi = \arccos(\cos(\alpha) \cos(\beta)), 
%	\label{eq_tilt}
%\end{equation}

% Tilt

Pitch $\alpha $ and roll $\beta$ of a USV are a function of the wavelength, wave amplitude, and vessel length and width \cite{sears2023mapping}. The tilt angle $\phi$ of a USV can then be determined:
\vspace{-2mm}
\begin{equation}
	\phi = \arccos(\cos(\alpha) \cos(\beta)), 
	\label{eq_tilt}
 \vspace{-1mm}
\end{equation}
where the tilt $\phi$ is defined as the angle between the direction of the gravity vector measured in the body frame $B$ and the gravity vector measured in the inertial world $W$ frame, see Fig. \ref{F1}. To mimic the effect of spatial-temporal waves on a USV, similar to \cite{sears2023mapping}, we model the tilt motion of our platform as a dynamic scalar field $\phi(\mathbf{q},t): \mathcal{Q} \times \mathbb{R}_{\geq0} \rightarrow \mathbb{R}$ where $\mathcal{Q}\subset \mathbb{R}^2$ is the spatial domain of interest and time $t \in \mathbb{R}_{\geq0}$. We define the map $f_w(\mathbf{q},t): \mathcal{Q} \times \mathbb{R}_{\geq0} \rightarrow \mathbb{R}_{\geq0}$ as $f_w(\mathbf{q},t) = (\phi(\mathbf{q},t))^2$. 

Defining the input $\mathbf{a} = [\mathbf{q}, t]$, we propose to learn the spatial temporal map $f_w(\mathbf{a})$ using GP regression. Given $N_d$ noisy observations $\mathcal{D} = \{ \mathbf{a}_j, \hat{\phi}^2_j \}^{N_d}_{j=1}$ where $\hat{\phi}^2_j$ is a measurement of $f_w(\mathbf{a})$ with zero mean Gaussian noise $\sigma_w^2$, the posterior mean and variance at a query point $\mathbf{a}^*$ conditioned on the observed data $\mathcal{D}$ are \cite{rasmussen2006gaussian}:
\vspace{-2mm}
\begin{equation}
	\mu(\mathbf{a^*}) = \mathbf{k}(\mathbf{a^*}) \mathbf{K}^{-1} \mathbf{\hat{\Phi}}, 
	 \label{eq_mean}
\end{equation}
\vspace{-8mm}
\begin{equation}
	\sigma^2(\mathbf{a^*}) = k(\mathbf{a^*}, \mathbf{a^*}) - \mathbf{k}(\mathbf{a^*}) \mathbf{K}^{-1} \mathbf{k}^\intercal(\mathbf{a^*}),
	\label{eq_var}
\end{equation}
where $\mathbf{\hat{\Phi}} =[\hat{\phi}^2_1, \hat{\phi}^2_2, ..., \hat{\phi}^2_{N_d}]^\intercal$ is the vector of observed function values, the covariance matrix has entries $\mathbf{K}_{(i,j)} = k(\mathbf{a}_i, \mathbf{a}_j), \quad i,j \in {1, ..., {N_d}}$, and $\mathbf{k}(\mathbf{a}) = [k(\mathbf{a^*}, \mathbf{a}_1), ..., k(\mathbf{a^*}, \mathbf{a}_{N_d})]$ is the vector of the covariances between the query point $\mathbf{a^*}$ and the observed data points in $\mathcal{D}$. The kernel selection could be the SE kernel from (\ref{eq_se}) or other common kernel options, including the Matern kernel \cite{rasmussen2006gaussian} or periodic kernel \cite{mackay1998introduction}.

%In this way, the USV acts as a `sensor' that can be used to model the wave map $f_W(\mathbf{x})$ as it moves in space and time. As the USV moves it gathers one data point every 0.1 seconds and replaces the oldest data in the set $\mathcal{D}$ with current observations. Accordingly, each second the GP is provided with ten new observations and is reconditioned on the most current $N$ noisy observations, $\mathcal{D} = \{ \mathbf{x}_i, \hat{\phi}^2_i \}^N_{i=1}$.

\subsection{Distributed MPC}
Similar to our previous work in \cite{stephenson2024distributed}, we propose a distributed MPC scheme to enable cooperative control of the multirotor and the tilting and moving platform whose heterogeneous dynamics are decoupled and nonlinear.  To do this, we augment a standard tracking MPC, see (\ref{eq_mpc}), for both a multirotor and the moving platform with artificial setpoint goals, defined as $\mathbf{x}_g^m$ for the multirotor and $\mathbf{x}_g^s$ for the platform. At each timestep of the multirotor MPC, we optimize for the multirotor sequence of states $\mathbf{x}^m_{0:N_m}$ and inputs $\mathbf{u}^m_{0:N_m-1}$ as well as the setpoint goal $\mathbf{x}_g^m$ by solving the constrained optimization problem:
\begin{equation}
\begin{aligned}
	\min_{\mathbf{x}^m_{0:N_m}, \mathbf{u}^m_{0:N_m-1}, \mathbf{x}^m_g} & \quad J^m(\mathbf{x}^m_{0:N_m}, \mathbf{u}^m_{0:N_m-1}, \mathbf{x}^m_g, \mathbf{x}^s_g) \\
\textrm{s.t.} \quad & \mathbf{x}^m_{k+1} = f^m(\mathbf{x}^m_k, \mathbf{u}^m_k) \quad \forall k \in \mathcal{K}^m \\ %= 0,..., N-1\\
  &\mathbf{x}^m_{k}, \mathbf{x}^m_{g} \in \mathcal{X}^m \quad \forall k \in \mathcal{K}^m  \\
  &\mathbf{u}^m_{k} \in \mathcal{U}^m \quad  \forall k \in \mathcal{K}^m  \\
  &\mathbf{x}^m_{0} = \mathbf{x}^m_{\text{init}},
\end{aligned}
\label{eq_mpc_multirotor}
\end{equation}
where $N_m \in \mathbb{Z}$ is the multirotor's MPC prediction horizon, $\mathcal{K}^m := \mathbb{Z} \cap [0, N_m-1]$, $f^m(\cdot)$ is used to denote the multirotor time-discretized dynamics in (\ref{eq_dyn}), $\mathcal{U}^m$ is the input space for multirotor system, $\mathcal{X}^m$ is the state space for multirotor system, such that $f^m : \mathcal{X}^m \times \mathcal{U}^m \rightarrow \mathcal{X}^m$, $\mathbf{x}^m_{\text{init}}$ is the current estimated multirotor state. The multirotor cost $J^m(\cdot)$ is designed to account for tracking, cooperation, and tilting costs. At each time step, we assume that the artificial platform goal $\mathbf{x}_g^s$ has been shared, and optimize (\ref{eq_mpc_multirotor}) for the sequence of optimal states and inputs as well as the optimized multirotor goal $\mathbf{x}_g^{m*}$. This optimized goal $\mathbf{x}_g^{m*}$ is then shared with the cooperative moving platform. 

Independently and asynchronously, the platform is controlled using its own onboard MPC. At each timestep of the platforms MPC, we optimize for the platform's sequence of states $\mathbf{x}^s_{0:N_s}$ and inputs $\mathbf{u}^s_{0:N_s-1}$ as well as the setpoint goal $\mathbf{x}_g^s$ by solving the constrained optimization problem:
\begin{equation}
\begin{aligned}
	\min_{\mathbf{x}^s_{0:N_s}, \mathbf{u}^s_{0:N_s-1}, \mathbf{x}^s_g} & \quad J^s(\mathbf{x}^s_{0:N_s}, \mathbf{u}^s_{0:N_s-1}, \mathbf{x}^s_g, \mathbf{x}^m_g) \\
\textrm{s.t.} \quad & \mathbf{x}^s_{k+1} = f^s(\mathbf{x}^s_k, \mathbf{u}^s_k) \quad \forall k \in \mathcal{K}^s \\ %= 0,..., N-1\\
  &\mathbf{x}^s_{k}, \mathbf{x}^s_{g} \in \mathcal{X}^s \quad \forall k \in \mathcal{K}^s  \\
  &\mathbf{u}^s_{k} \in \mathcal{U}^s \quad  \forall k \in \mathcal{K}^s  \\
  &\mathbf{x}^s_{0} = \mathbf{x}^s_{\text{init}},
\end{aligned}
\label{eq_mpc_platform}
\end{equation}
where $N_s \in \mathbb{Z}$ is the platform's MPC prediction horizon, $\mathcal{K}^s := \mathbb{Z} \cap [0, N_s-1]$, $f^s(\cdot)$ is used to denote the platform time-discretized dynamics in (\ref{eq_dyn}), $\mathcal{U}^s$ is the input space for platform system, $\mathcal{X}^s$ is the state space for platform system, such that $f^s : \mathcal{X}^s \times \mathcal{U}^s \rightarrow \mathcal{X}^s$, $\mathbf{x}^s_{\text{init}}$ is the current estimated platform state. The platform cost $J^s(\cdot)$ is designed to account for tracking, cooperation, and tilting costs. At each time step, we assume that the artificial multirotor goal $\mathbf{x}_g^m$ has been shared, and optimize (\ref{eq_mpc_platform}) for the sequence of optimal states and inputs as well as the optimized platform goal $\mathbf{x}_g^{s*}$. This optimized goal $\mathbf{x}_g^{s*}$ is then shared with the multirotor.

%the cooperative landing on a tilting platform sequential distributed MPC scheme allowing for cooperative control of multi-agent systems with dynamically decoupled heterogeneous nonlinear agents.

%(\ref{eq_mpc}) represents a standard tracking MPC that is applied to USV in \cite{fossen2011handbook} and to UAV in \cite{nan2022nonlinear}. 

%We augment this tracking MPC scheme by leveraging a sequential distributed MPC scheme allowing for cooperative control of multi-agent systems with dynamically decoupled heterogeneous nonlinear agents. To do this we apply an artificial goal in the tracking MPC for both the USV and UAV. The USV MPC optimizes for the artificial USV position goal $\mathbf{p}^s_{G}$ at each time step, where the last artificial multirotor goal $\mathbf{p}^m_{G}$ is communicated to the USV. We solve the optimization as:
%\begin{equation}
%	\min_{\mathbf{x}^s_{0:N_S}, \mathbf{u}^s_{0:N_S-1}, \mathbf{p}^s_G} & \quad J_S(\mathbf{x}^s_{0:N_S}, \mathbf{u}^s_{0:N_S-1}, \mathbf{p}^s_G, \mathbf{p}^m_G) \\
%\label{eq_mpc_usv}
%\end{equation}

We design the platform cost $J^s(\cdot)$ in (\ref{eq_mpc_platform}) to account for three cost terms, a tracking cost $J^s_{\textrm{track}}(\cdot)$, a cooperative cost $J^s_{\textrm{coop}}(\cdot)$ and a tilt cost $J^s_{\textrm{tilt}}(\cdot)$ as:
\begin{equation}
    J^s(\cdot) = J^s_{\textrm{track}}(\cdot) + J^s_{\textrm{coop}}(\cdot) + J^s_{\textrm{tilt}}(\cdot),
\end{equation}
where we design the tracking cost $J^s_{\textrm{track}}(\cdot)$ as a standard quadratic cost (\ref{eq_cost_track}). Specifically,  $J^s_{\textrm{track}}(\cdot) = \sum_{k=1}^{N_s} (\mathbf{{x}}^s_k - \mathbf{x}^s_g)^\intercal\mathbf{Q}_s( \mathbf{{x}}^s_k - \mathbf{x}^s_g) + {{\mathbf{u}^s}^\intercal_{k-1}} \mathbf{R}_s \mathbf{u}^s_{k-1}$ where $\mathbf{Q}_s \succeq 0$ and $\mathbf{R}_s \succ 0$ are selected weights on the position error (between the platform's state and its goal) and control effort, respectively. The cooperative cost $J^s_{\textrm{coop}}(\cdot)$ penalizes the deviation of the artificial goals to encourage finding a cooperative landing location:
\begin{equation}
    J^s_{\textrm{coop}}(\cdot) = (\mathbf{{x}}^s_g - \mathbf{x}^m_g)^\intercal\mathbf{W}_s (\mathbf{{x}}^s_g - \mathbf{x}^m_g),
\end{equation}
where $\mathbf{W}_s \succcurlyeq 0$ is a positive semidefinite matrix. The tilt cost $J^s_{\textrm{tilt}}(\cdot)$ is designed to encourage the optimization of a cooperative goal location where the average tilt motion of the platform is minimized. We proposed leveraging the spatial-temporal map $f_w(\mathbf{a})$ learned as GP, see Sec. \ref{meth_gp}. 

We determine the x-y spatial position of the goal state as $\mathbf{q}^* = \mathbf{C} \mathbf{x}^s_g$ where $\mathbf{q}^* \in \mathbb{R}^2$ and $\mathbf{C}$ is a matrix that selects out the position from the goal state. We define the query inputs of the GP as $\mathbf{a}_j^* = [\mathbf{q}^*, t^*_j]$ where $j = 0, 1, ... , N_w, j\in \mathbb{Z}$ and $N_w$ is the number of sample points of the GP. The query times $t^*_j = \frac{T_w}{N_w} j$ where $T_w$ is the temporal period of the tilt. We design the tilt cost as:
\vspace{-2mm}
\begin{equation}
    J^s_{\textrm{tilt}}(\cdot) = \lambda_w \sum_{j=0}^{N_w} \mu(\mathbf{a}_j^*) + \lambda_v \sum_{j=0}^{N_w} \sigma^2(\mathbf{a}_j^*),
    \label{eq_tilt_s}
    \vspace{-1mm}
\end{equation}
where $\lambda_w > 0$, $\lambda_v > 0$ are positive weights, $\mu(\mathbf{a}_j^*)$ is the posterior mean (\ref{eq_mean}) and $\sigma^2(\mathbf{a}_j^*)$ is posterior covariance (\ref{eq_var}). This cost encourages the computation of a platform goal location where a small tilt is likely while weighting this with uncertainty in the tilt model.  

We similarly design the multirotor cost $J^m(\cdot)$ in (\ref{eq_mpc_multirotor}) to account for three cost terms, a tracking cost $J^m_{\textrm{track}}(\cdot)$, a cooperative cost $J^m_{\textrm{coop}}(\cdot)$ and a tilt cost $J^m_{\textrm{tilt}}(\cdot)$ as:
\begin{equation}
    J^m(\cdot) = J^m_{\textrm{track}}(\cdot) + J^m_{\textrm{coop}}(\cdot) + J^m_{\textrm{tilt}}(\cdot),
\end{equation}
where we design the tracking cost $J^m_{\textrm{track}}(\cdot)$ as a standard quadratic cost (\ref{eq_cost_track}). Specifically,  $J^m_{\textrm{track}}(\cdot) = \sum_{k=1}^{N_m} (\mathbf{{x}}^m_k - \mathbf{x}^m_g)^\intercal\mathbf{Q}_m( \mathbf{{x}}^m_k - \mathbf{x}^m_g) + {{\mathbf{u}^m}^\intercal_{k-1}} \mathbf{R}_m \mathbf{u}^m_{k-1}$ where $\mathbf{Q}_m \succeq 0$ and $\mathbf{R}_m \succ 0$ weigh the position error (between the multirotor's state and its goal) and control effort, respectively. The cooperative cost $J^m_{\textrm{coop}}(\cdot)$ penalizes the deviation of the artificial goals to encourage finding a cooperative landing location:
\begin{equation}
    J^m_{\textrm{coop}}(\cdot) = (\mathbf{{x}}^m_g - \mathbf{x}^s_g)^\intercal\mathbf{W}_m (\mathbf{{x}}^m_g - \mathbf{x}^s_g),
\end{equation}
where $\mathbf{W}_m \succcurlyeq 0$ is a positive semidefinite matrix. The tilt cost $J^m_{\textrm{tilt}}(\cdot)$ is designed to encourage the optimization of the timing of the landing when the tilt is low. We leverage the spatial-temporal map $f_w(\mathbf{a})$ learned as GP. See Sec. \ref{meth_gp}. 

To do this, we first design a landing function, similar to  \cite{gupta2022landing}, as:
\vspace{-2mm}
\begin{equation}
    h(\mathbf{e}_{g}) = \begin{cases}
(1 + \text{exp}(-\frac{\mathbf{e}_{g} - h_d}{-0.15}))^{-1} &\text{if $\mathbf{e}_{g} \geq 0.16$}\\
(1 + \text{exp}(\frac{\mathbf{e}_{g} - h_d}{-0.01}))^{-1} &\text{otherwise,}
\end{cases}
\label{eq_land_fcn}
\end{equation}
where $h_d$ is a holding height, $\mathbf{e}_g$ is the error in the altitudes of the goals, $\mathbf{e}_g = \mathbf{D} (\mathbf{x}_g^m - \mathbf{x}_g^s)$ and $\mathbf{D}$ is a matrix that selects out the goals altitude error. We define query points $\mathbf{a}^*_k = [\mathbf{q}^*, t^*_k]$ where $k = 0, 1, ..., N_m$ and $t^*_k = t_0 + \delta_m k$, $t_0$ is the current time, $\delta_m$ is the time-discretization for the multirotor model in (\ref{eq_mpc_multirotor}). 

Similar to \cite{gupta2022landing}, we only activate the tilt cost $J^m_{\textrm{tilt}}(\cdot)$ when two conditions hold. Firstly, the platform state has converged to its goal, $| \mathbf{x}^s - \mathbf{x}^s_g | \leq \epsilon_p$ for some small $\epsilon_p \geq 0$. When this is achieved, the platform communicates to the multirotor that ``landing is possible". The second condition is that multirotor state has converged to its goal, $| \mathbf{x}^m - \mathbf{x}^m_g| \leq \epsilon_q$. When these conditions hold $J_{\textrm{tilt}}^m(\cdot)$ is activated as:
\vspace{-2mm}
\begin{equation}
    J_{\textrm{tilt}}^m(\cdot) = \lambda_u \sum_{k=1}^{N_m} h(\mathbf{e}_{g}) \mu(\mathbf{a}_k),
    \label{eq_tilt_M}
\end{equation}
where $\lambda_u > 0$ is a positive weight and    $\mu(\mathbf{a}_k)$ is the posterior mean (\ref{eq_mean}). The tilt costs (\ref{eq_tilt_s}) and  (\ref{eq_tilt_M}) encourage cooperation of the multirotor and platform to a location and time of landing when the tilt of the platform is minimized.    

\begin{table}[t]
\centering
\vspace{5mm}
\caption{Specifications of the Tilting Landing Platform}
\renewcommand{\arraystretch}{1.5}  % Adjusts the height of rows
\begin{tabular}{ll}
\hline
\multicolumn{1}{l}{\textbf{Specification}} & \multicolumn{1}{l}{\textbf{Value/Description}} \\ \hline
%\rule{0pt}{3ex}
Platform Size         & Square, 0.5 m across
\\
Surface Material      & 0.5 mm acrylic sheet               
\\
Maximum UAV Mass      & 500 g
\\
Platform Tilt Range   & -60° to 60° 
\\
Angular Resolution      & 0.5° 
\\
Maximum Angular Rate  & 135°/s                             
\\
%Maximum Oscillation Frequency & 0.56 Hz (at 60° tilt)      \\ \hline   % I'm leaving this commented because this can easily be deduced using the tilt range and the maximum angular rate
Servo Motor Torque      & 3.4 Nm                            
\\
Servos Per Axis          & 2                              
\\
Degrees of Freedom (DOF)          & 2 (pitch, roll)         
\\
%\multirow{1}{*}{Torque from Quadcopter Weight} & 1.3 Nm                     \\ \hline
%\multirow{1}{*}{Inertia Torque (Platform)} & 0.1 Nm                         \\ \hline
Moment of Inertia I\(_{xx}\), I\(_{yy}\)  & 1.4 × $10^7$ g·mm$^2$  
\\ \hline
%\multirow{1}{*}{Safety Factor}         & 4                                \\ \hline
%\multirow{1}{*}{Servo Load (per axis)} & 1.4 Nm load distributed between two 3.4 Nm servos    \\ \hline
%\multirow{1}{*}{Rotation Mechanism}    & 4-bar linkage                   \\ \hline
%\multirow{1}{*}{Friction Assumption}   & Ignored (calculated without friction)                \\ \hline
\end{tabular}
\label{platform_specs}
 \vspace{-5mm}
\end{table}

\subsection{Tilting Platform} % Will's Section
Our tilting platform, see Fig. \ref{husky}, is designed as a testbed for indoor experimental validation of our proposed distributed MPC for landing on a USV. We design and build a custom tilting platform with two degrees of freedom (roll, pitch), which can be affixed to the deck of a differential-drive UGV, the ClearPath Robotics Husky. We selected compatibility with a differential-drive UGV to replicate the motion of a broader range of USVs, including those with differential drive and conventional rudder steering. Our platform is relatively low-cost and modular, enabling rapid prototyping and testing of landing schemes for USVs.

%was developed to be a low cost, modular, open source solution to emulating the motion of a USV in heavy waves. A custom-designed tilting platform with two degrees of freedom (roll, pitch) is affixed to the deck of a pre-existing differential-drive UGV. Vertical translation was deemed to be unnecessary because \textbf{XXXXXXXXXXX}. We designed our platform around a differential-drive UGV to better replicate the motion of a wider range of USVs, including those with differential drive and conventional rudder steering. This is thanks to the ability of differential drive to pivot around its central axis, unlike the Ackermann-steering configuration. The UGV we selected is the ClearPath Robotics Husky. Please note that we do not consider the UGV itself to be low-cost, only the landing platform affixed to it.

The acrylic platform is designed to support the weight of a $500$ g multirotor landing at the platform's rim. At a tilt $\phi = 0$, the landing pad is parallel to the ground, see Fig. \ref{F1}. The platform is designed to achieve a maximum roll and pitch of $60^\circ$ and a maximum angular rate of $135^\circ/s$. See Table \ref{platform_specs} for key platform specifications.

\subsubsection{Physical Configuration} 
The platform is centered around an elevated ball-and-socket joint (see Fig. \ref{husky_section}), enabling a $60^\circ$ tilt in both roll and pitch. Each axis features two servo-driven four-bar linkages that apply vertical forces at four off-center points to adjust the landing pad’s orientation. These four-bar linkages are radially spaced at $90^\circ$ intervals, ensuring that one servo on each axis is always pushing up while the other pulls down. Universal joints on the linkages allow for smooth adaptation to changes in pad orientation across multiple axes, with their centers aligned in a plane parallel and vertically offset from the landing pad. This configuration creates symmetrical movement about the ball joint, allowing servos on the same axis to receive identical angle commands with opposite signs for control.
%The ball-and-socket joint shares this alignment, ensuring full range of motion without singularities.
%Excluding fasteners, the universal joints, and the servo motors, all structural components are either lasercut, 3D printed, or off-the-shelf aluminum extrusion. The landing surface is lasercut 5 mm acrylic; the linkages and motor mounts, fused deposition modeling (FDM) 3D printed; and the ball and socket joint, stereolithography (SLA) 3D printed. These manufacturing choices were made to keep the cost of the testbed low and allow for repairs to easily be made if necessary. The linkages and motor mounts were made using PLA plastic and the FDM manufacturing process due to the need for inexpensive parts with moderate specific strength and a minimal need for surface finish quality. The ball and socket joint was made using resin and the SLA manufacturing process due to the need for inexpensive parts with high surface finish quality and minimal specific strength requirements. The landing pad features a lasercut hexagonal pattern which is intended to remove unnecessary material from the landing pad to reduce its moment of inertia.
Most components are lasercut, 3D printed, or off-the-shelf aluminum extrusions to keep costs low and allow easy repairs. The landing surface is 5 mm lasercut acrylic; the linkages and motor mounts are 3D printed with PLA for strength and cost-efficiency, while the ball-and-socket joint is 3D printed using SLA for high surface finish. A lasercut hexagonal pattern in the landing pad reduces its moment of inertia by removing unnecessary material.

\subsubsection{Electronics and Software Specifications}
We use an Arduino Uno for the PWM control of our tilting platform. We use a Raspberry Pi onboard the UGV, with Ubuntu 20.04 and ROS 1 Noetic, to run the vehicle's MPC (\ref{eq_mpc_platform}). The Raspberry Pi computes the tilt for both axes and sends them to an Arduino Uno via USB serial connection. The Arduino computes the angles about the platform's neutral position and sends opposite PWM signals to servos on the same axis.

\begin{figure}[t]
 \centering
 \subfigure[Experiment 1 - Calm Waves]{
      \centering
      \label{box_a}
      \includegraphics[width=0.47\textwidth,trim={0cm 0cm 0cm 0cm},clip]{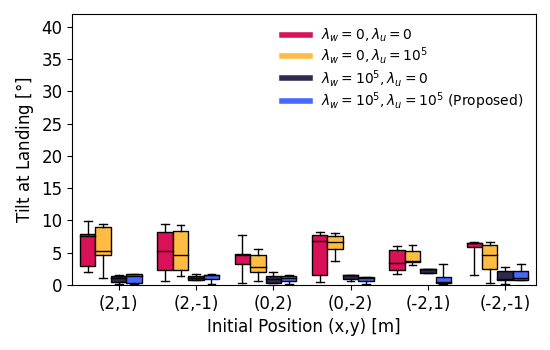}
		} \\
 \subfigure[Experiment 2 - Harsh Waves]{
      \centering
      \label{box_b}
      \includegraphics[width=0.47\textwidth,trim={0cm 0cm 0cm 0cm},clip]{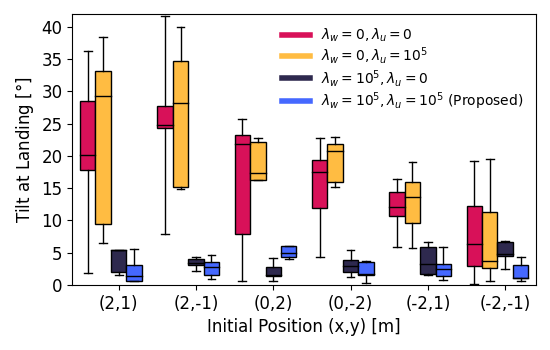}
		}
 \caption{Experiment 1 and 2: We visualize the tilt of the platform at landing in (a) calm wave conditions and (b) harsh wave conditions. In both experiments, our proposed approach (blue) achieves a lower landing tilt at landing over the purely cooperative strategy (red). Our proposed distributed MPC scheme can locate a low-tilt landing location from all six initial platform positions.
 }
 \label{box_plots_1}
 \vspace{-5mm}
 \end{figure}

 \begin{figure*}[ht]
 \centering
 \subfigure[Landing Location]{
      \centering
      \label{ex2_3}
      \includegraphics[width=0.3\textwidth]{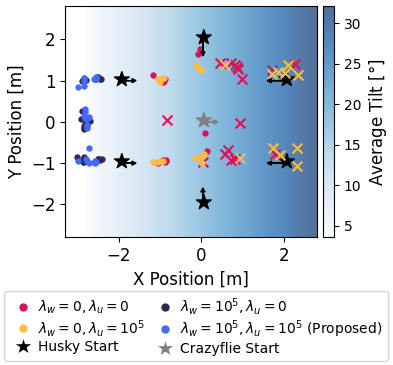}
		} 
 \subfigure[Testbed Platform Landing]{
      \centering
      \label{ex2_2}
      \includegraphics[width=0.3\textwidth]{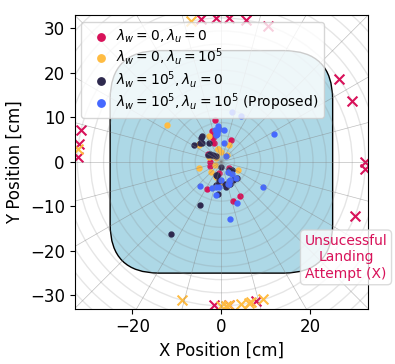}
		} 
\subfigure[Energy Expenditure]{
      \centering
      \label{ex2_4}
      \includegraphics[width=0.3\textwidth]{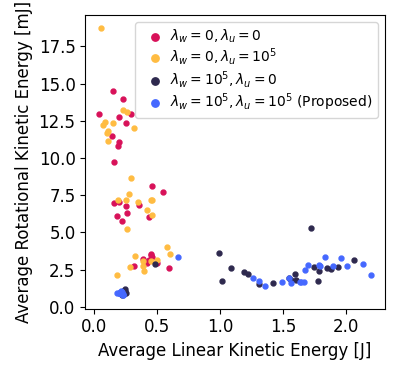}
		} 
 \caption{Experiment 2 visualization of (a) final $x$-$y$ landing locations, (b) landing locations on the platform, and (c) the average linear vs rotational kinetic energy of the tilting platform. Failed landings are represented by an $\times$.}
 \label{ex2}
  \vspace{-5mm}
\end{figure*}

\begin{figure}[t]
\centering
\includegraphics[width=0.47\textwidth,trim={0cm 0cm 0cm 0cm},clip]{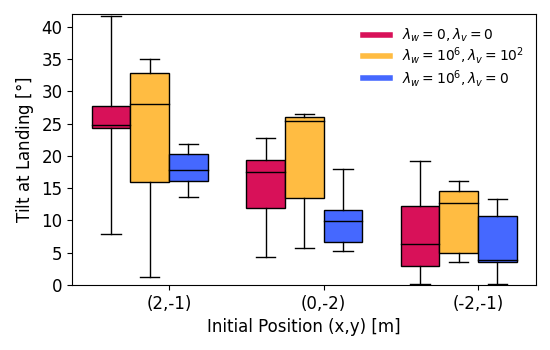}
 \caption{Experiment 3: We learn the wave model from Experiment 2 using a GP. We visualize the tilt of the platform at the landing using pure cooperation (red), a mean cost and high variance cost (yellow), and only a mean cost (blue).}
 \label{box_ex3}
 %\vspace{-5mm}
\end{figure}
\section{EXPERIMENTAL RESULTS}  %2 pages

We evaluate our proposed distributed MPC scheme for UAV-USV cooperative landing in three sets of indoor experiments using our developed testbed platform from Fig. \ref{husky}. Our experimental set-up comprises a Bitcraze Crazyflie 2.1 UAV and a ClearPath Robotics Husky mounted with our tilting platform. The UAV MPC (\ref{eq_mpc_multirotor}) is run off-board on a Thinkpad X1 Carbon with Intel Core i7-1270P Processor. The MPC runs at a frequency of 50Hz and transmits control inputs to the Crazyflie via long range Crazyradio USB. The Raspberry Pi onboard the Husky runs its MPC (equivalent to (\ref{eq_mpc_platform}) for UGVs) at 10Hz. The two vehicles communicate their goals using ROS topics on a local WiFi network and receive their own pose feedback at 240 Hz from a VICON motion capture system (MCS). The Crazyflie 2.1 starts at $x$-$y$ position $(0,0)$ and an altitude of 1.3 meters.

\subsection{Experiments 1 and 2: Prescribed Wave Model}
In Experiments 1 and 2, we assume no uncertainty in the tilt model. Consequently, $\lambda_v = 0$ in (\ref{eq_tilt_s}) and the mean is replaced with the map value, $\mu(\mathbf{a}^*) = f_w(\mathbf{a}^*)$ in (\ref{eq_tilt_s}) and (\ref{eq_tilt_M}). We compare 4 MPC costs (or strategies) initializing the tilting platform at six different starting positions. For each case, we perform five trials. The strategies are 1) pure cooperation (red) where no tilt cost $J^{\text{tilt}}(\cdot)$ for either vehicle is considered, UAV tilt cost only (yellow) with $\lambda_u=10^5$ and $\lambda_w=0$, USV tilt cost only (black) with $\lambda_u=0$ and $\lambda_w=10^5$, and our proposed approach (blue) $\lambda_u=10^5$ and $\lambda_w=10^5$. Our tilt model takes the form $\theta = A (\mathbf{q}_x + b) \sin(\pi(T + t + \mathbf{q}_x))$, where $\mathbf{q}_x$ is the $x$-direction position, $b = 3.5$, $T=5$. In Experiment 1, we simulate `calmer' wave conditions using our platform by setting $A= 2.3$. In Experiment 2, we simulate `harsher' wave conditions using our platform by setting $A = 8$. \textit{Experiment 1 - Calm Waves:} All approaches achieve 100\% landing success rate. Although our proposed approach (blue) achieves a lower tilt (71-83\% reduction) at landing from all initial positions over a purely cooperative strategy (red), both approaches result in landing at a relatively low tilt in the `calm' wave conditions, see Fig. \ref{box_a}. Consequently, we observe no significant safety benefit to exploiting the tilt model in `calm' wave conditions. \textit{Experiment 2 - Harsh Waves:} In Fig. \ref{box_b}, we observe that similar to Experiment 1, our proposed approach (blue) reduces the tilt angle of the platform at landing by between 68-89\% over a purely cooperative strategy (red). Our approach (blue) has a 53\% increase in landing success rate over the cooperative strategy (red). Fig. \ref{ex2_3} shows the landing location of each trial, where $\times$ is used to mark an associated unsuccessful landing. We observe that successful landing is not achieved at locations with higher average tilt. These high-tilt locations are optimized for using the cooperation strategy (red) or the UAV tilt cost only strategy (yellow) because these coordination approaches neglect the spatial component of the tilt model. The UAV tilt cost only strategy (yellow) still achieves a 17\% increase in the landing success rate over the pure cooperation strategy (red) - see Fig. \ref{ex2_2}. In Fig. \ref{ex2_4}, we observe the relationship between the average rotational kinetic energy of the platform and the average linear kinetic energy expended by the UGV for each trial. Our proposed approach (blue) achieves a lower average rotational kinetic energy at the cost of a higher average linear kinetic energy. This results from the platform traveling further to locate a safe landing location.

\subsection{Experiment 3: Learned Wave Model} %GP online learning
We learn the model from Experiment 2 as a GP with 50 data points, sampled using Latin hypercubic sampling \cite{McKay1979239} with $\mathbf{q}_x \in [0.5, -0.5]$. We use a standard SE kernel (\ref{eq_se}). We compare pure cooperation (red) with weighting a high posterior covariance cost (yellow), i.e., $\lambda_w=10^6$ and $\lambda_v=10^2$, and no posterior covariance cost (blue), $\lambda_w=10^6$ and $\lambda_v = 0$. We consider three starting locations for the platform and perform five trials for each case. Weighting the posterior mean cost (blue) facilitates landing at a lower tilt - see Fig. \ref{box_ex3}. This approach reduces the tilt at landing by 23-32\% and results in a 53\% increase in the landing success rate over pure cooperation (red). However, a high additional weighting of the covariance (uncertainty in the model) may lead to a higher tilt at landing due to the location of the data and associated uncertainty. 

\section{CONCLUSION}
This paper uses a custom-built testbed to present indoor experimental validation of our novel learning-based distributed MPC framework for cooperative UAV-USV landing. We demonstrate a significant improvement in landing success rate over alternative strategies. Future work will extend our experimental results to outdoor field trials landing on a USV in waves.

\bibliography{References.bib}
\bibliographystyle{ieeetr}

\end{document}